\pdfoutput=1

\documentclass[11pt]{article}

\usepackage[final]{nlp4MusA}

\usepackage{times}
\usepackage{latexsym}
\usepackage{enumitem}
\usepackage{textcomp}
\usepackage{subcaption}
\usepackage{paralist}

\usepackage[T1]{fontenc}

\usepackage[utf8]{inputenc}

\usepackage{microtype}

\usepackage{inconsolata}

\usepackage{graphicx}

\title{Harmonic Reasoning in Large Language Models}

\author{Anna Kruspe \\
  Munich University of Applied Sciences \\
  Lothstr. 64, 80335 Munich \\ 
  Germany \\
  \texttt{anna.kruspe@hm.edu} }

\begin{document}
\maketitle
\begin{abstract}

Large Language Models (LLMs) are becoming very popular and are used for many different purposes, including creative tasks in the arts. However, these models sometimes have trouble with specific reasoning tasks, especially those that involve logical thinking and counting. This paper looks at how well LLMs understand and reason when dealing with musical tasks like figuring out notes from intervals and identifying chords and scales. We tested GPT-3.5 and GPT-4o to see how they handle these tasks. Our results show that while LLMs do well with note intervals, they struggle with more complicated tasks like recognizing chords and scales. This points out clear limits in current LLM abilities and shows where we need to make them better, which could help improve how they think and work in both artistic and other complex areas. We also provide an automatically generated benchmark data set for the described tasks.
\end{abstract}

\section{Introduction}

Generative AI, particularly Large Language Models (LLMs), is increasingly utilized across a range of applications, extending beyond textual outputs to include artistic creations such as image generation, text composition, and music. This proliferation of applications has sparked interest in cross-modality, a growing field of research exploring the transfer and application of generative models across different sensory domains. Among these, musical generation tools like Suno\footnote{\url{https://suno.com/}} and Udio\footnote{\url{https://www.udio.com}} have demonstrated an implicit understanding of musical rules, raising questions about whether LLMs, which have been exposed to vast amounts of music-related literature during training, exhibit a similar grasp of musical concepts.

Western music is characterized by a rigorously structured system encompassing rhythm, harmony, and other elements fundamental to both composition and music analysis. This structured nature of music presents a unique challenge: Do LLMs truly understand the underlying rules of Western musical harmony, or are they merely replicating patterns observed in data? This question aligns with the broader inquiry into the reasoning capabilities of LLMs — whether these models can leverage facts learned during training to solve novel, unseen tasks. \cite{yuan2024chatmusician} introduces the concept of ``Music Reasoning'', defined as ``the ability to estimate the varying harmonies, keys, rhythms, and other musical elements that are not explicitly annotated in a piece of music and are significant for music themes, progression, and styles''.

There is some evidence suggesting that larger models may develop emergent abilities, allowing them to perform better on tasks requiring deep, structural understanding and reasoning. This paper aims to investigate these capabilities specifically within the realm of musical harmony. Our research is divided into two main experiments: 1. applying intervals to musical notes, and 2. recognizing chords and scales, which requires not only knowledge of these musical components, but also the ability to apply this knowledge to specific problems.

The paper is structured as follows: In the next section, we discuss related work about reasoning in LLMs, and especially in the music domain. In section \ref{sec:methodology}, we introduce our methodology for generating test questions for the interval and chords/scales tasks. Sections \ref{sec:res_intervals} and \ref{sec:res_cs} present the results. Finally, sections \ref{sec:conclusion} and \ref{sec:future} discuss the outcomes in detail and give suggestions for future work.

\section{Related work}\label{sec:related}
The integration of reasoning capabilities into large language models (LLMs) represents a significant frontier in artificial intelligence research, focusing on enhancing the models' ability to understand, interpret, and solve complex tasks. Across several key studies, researchers have employed various methodologies to push the boundaries of what LLMs can achieve in reasoning tasks.

Studies like \cite{reasoning_survey_2022} detail the development of sophisticated prompting techniques, such as chain-of-thought prompting, which guides LLMs to articulate their reasoning steps explicitly, thereby enhancing their interpretability and reliability.

In mathematical domains, LLMs confront complex problems that require not only numerical computations but also deep logical processing, as discussed in \cite{math_reasoning_2024}, e.g. theorem proving and complex problem solving in geometry. The application of LLMs to puzzle solving, reviewed in \cite{puzzle_solving_2022}, demonstrates their potential in creatively and logically applying learned knowledge. 

Despite significant advancements, the gap between current LLMs and human-like reasoning remains substantial. Efforts to bridge this gap involve refining training paradigms, developing richer datasets, and designing more complex model architectures. These efforts aim to enhance the models' abilities to handle nuanced understanding and multi-step logical deductions, as emphasized across the surveyed literature \cite{towards_reasoning_2022}.

Expanding into the music domain, LLMs are being explored for their capability to reason and create within the realm of music generation and theory. \cite{songcomposer_2024} investigates how LLMs can be adapted to generate music by understanding and applying complex musical structures and emotional cues, showing that LLMs can not only process but ``think'' in terms of music. Projects like ChatMusician \cite{yuan2024chatmusician}
serve to further test and enhance these capabilities, focusing on how well models can engage with music theory and respond creatively to musical queries.

\section{Methodology}\label{sec:methodology}
In this section, we will describe our experimental design. All generated questions were tested on GPT-4o. The interval experiments were also run on GPT-3.5 for comparison. Each experiment was performed three times to account for randomness in the models. Evaluation then consisted of comparing each result to the expected one to obtain accuracy, accounting for enharmonic variants.

\subsection{Interval problems}
To investigate the capability of Large Language Models (LLMs) in processing and understanding musical intervals, we utilized the \texttt{music21}\footnote{\url{https://pypi.org/project/music21/}} Python library to automatically construct a series of problems. These problems are formulated to challenge the LLM's ability to determine and name intervals based on the question format, ``What is a \textit{<interval>} up/down from \textit{<note>}?''. Intervals were limited to not be greater than an octave.

We introduced various configurations to increase the complexity of the task:
\begin{description}[style=unboxed, leftmargin=0cm, labelwidth=0cm, itemsep=0pt, parsep=0pt, partopsep=0pt, topsep=0pt]    \item[Direction:] Only upward intervals vs. upward and downward intervals 
    \item[Octave limitation:] Intervals remaining within the same octave (c to b) vs. intervals transcending octaves
    \item[Accidentals:] Sharps only vs. sharps and flats
\end{description}
We hypothesize that in each case, the first variant is more commonly seen in textbooks, and therefore easier to solve for LLMs presumably trained on such material. (The pitch classes Bb and Eb may be an exception for accidentals).

A comprehensive dataset comprising 500 questions for each configuration combination was compiled, ensuring a robust evaluation of the LLM's ability across different musical scenarios. Prior to testing, the LLM was briefed to expect questions about musical intervals and was instructed to format its responses in a table to standardize output.

\subsection{Chords and Scales Problems}

In the domain of chords and scales, the focus of the experiments was on the LLM’s ability to identify and name various types of chords and scales. Using the \texttt{music21} Python library, chords and scales were randomly constructed to create a diverse set of musical problems. The used chord and scale types are given in Table \ref{tab:cs} (appendix); definitions were used as in \texttt{music21}.

The evaluation was structured into two main experiments: In the first, the types of chords and scales were explicitly given to the LLM before it was asked to identify them. Interestingly, it was observed that the provided types were quickly ``forgotten'' or disregarded by the model in subsequent tasks, indicating a potential limitation in its short-term recall or application of explicitly given information. In the second experiment, the types were not provided, increasing the difficulty level and requiring the LLM to rely solely on its training and inherent understanding of musical structures.

Four configurations were tested for both chords and scales:
\begin{description}[style=unboxed, leftmargin=0cm, labelwidth=0cm, itemsep=0pt, parsep=0pt, partopsep=0pt, topsep=0pt]
    \item[Original order:] The most common presentation for each chord and scale, i.e. standard inversion. Using all chord and scale types and all twelve keys, this resulted in 108 test cases for the chords and 156 case for scales.
    \item[Original order with enharmonically exchanged notes:] Same as before, but some notes were randomly replaced with their enharmonic equivalents. Same amount of test cases.
    \item[Random permutations:] Chords and scale notes were given in random order(=inversions) with the original note names. Three permutations were created per chord and scale, resulting in 324 chord test cases and 468 scale test cases.
    \item[Random permutations with enharmonically exchanged notes:] Same as before, but once again, random notes are exchanged enharmonically.
\end{description}




\section{Interval Results}\label{sec:res_intervals}
\begin{figure}
    \centering
    \includegraphics[width=1\linewidth]{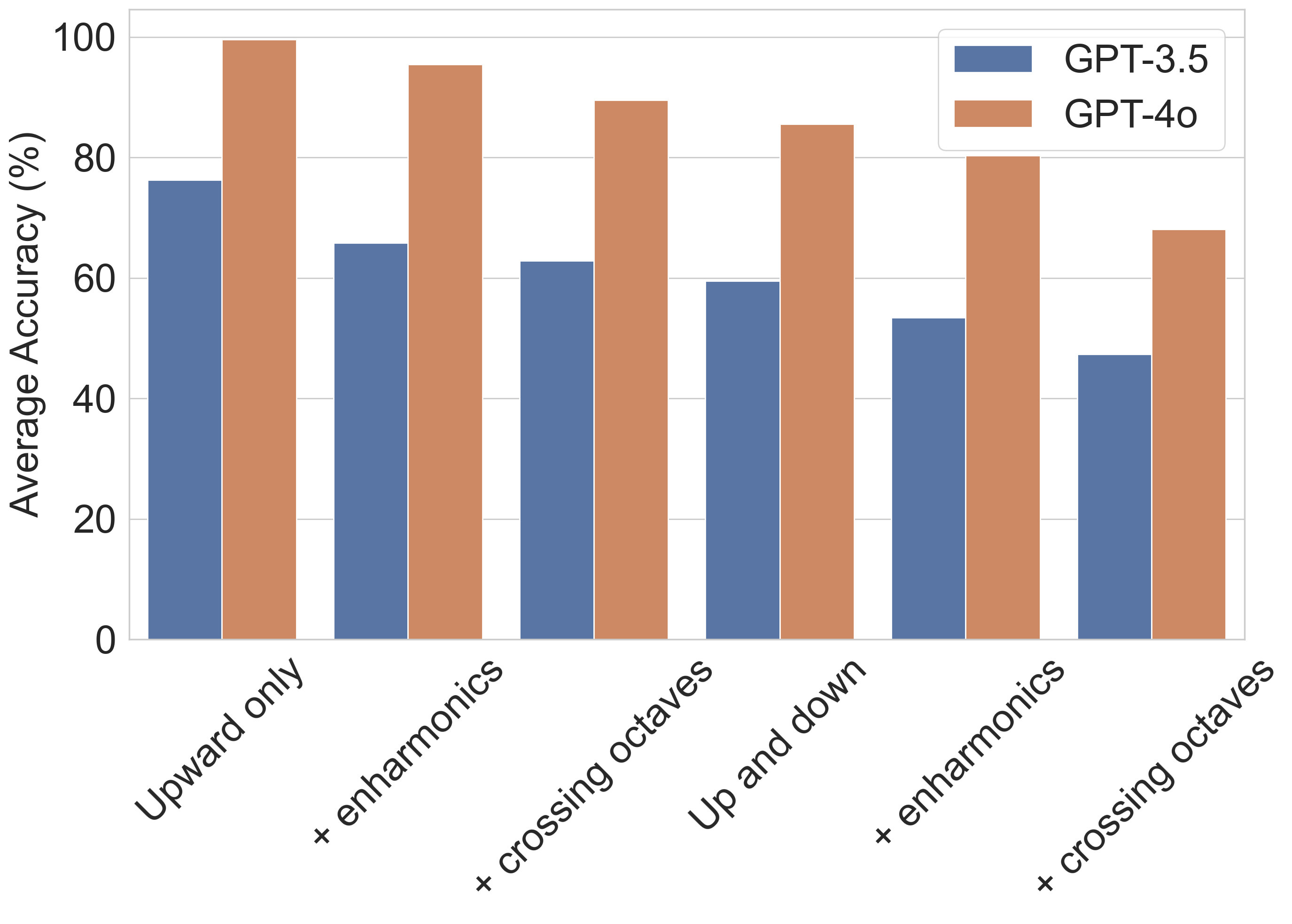}
    \caption{Results of the interval experiments, starting with upward intervals with sharps and within the same octave only, and then increasing difficulty of the task.}
    \label{fig:res_intervals}
\end{figure}

The results of the interval experiments are shown in Figure \ref{fig:res_intervals} for all described configurations, tested on GPT-3.5 and GPT-4o.

The results indicate that all the tested configurations influence the LLM's performance, with the direction of interval movement showing the strongest impact. This suggests a challenge in the LLM's ability to generalize from upward to both upward and downward interval calculations, which is less common in textbook material but critical for a robust understanding of musical intervals.

The more state-of-the-art model GPT-4o displayed markedly better performance, reinforcing the notion that both model size and the extent of training data significantly affect outcomes. This observation leads to a question: does the improvement signify better reasoning or merely a reflection of exposure to more data? For instance, the most advanced model nearly perfected the simplest task, achieving close to 100\% accuracy on upward intervals within a single octave. However, this accuracy declined to less than 50\% on the most challenging configurations involving downward movements and inclusion of both sharps and flats across multiple octaves when using GPT-3.5.

The implications of these results align with other findings in literature, suggesting that while LLMs can memorize and reproduce patterns seen during training, their logical reasoning capabilities, particularly in applying learned knowledge to novel situations, remain limited. The ability to memorize does not equate to a genuine understanding or logical processing, a distinction that becomes apparent in the more complex tasks where the LLM struggles to apply a robust logical framework to solve problems beyond straightforward recall.

\section{Chords and Scales results}\label{sec:res_cs}

\begin{figure*}
    \centering
    \begin{subfigure}[b]{.45\linewidth}
        \includegraphics[width=\linewidth]{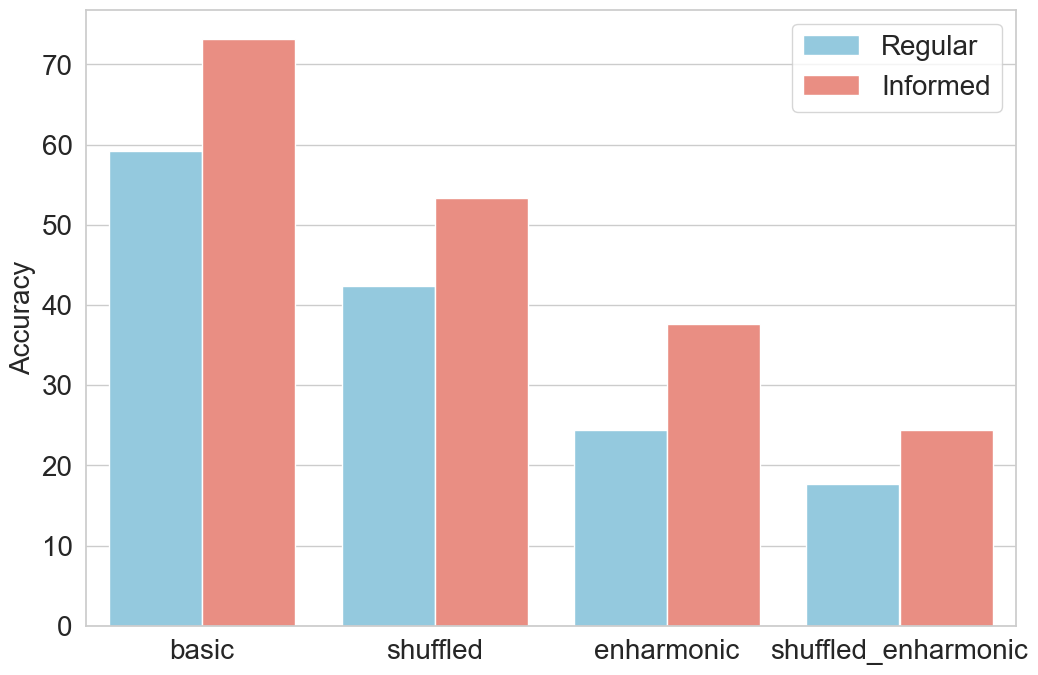}
        \caption{Chords}
        \label{fig:chords}
    \end{subfigure}
    \hfill
    \begin{subfigure}[b]{0.45\linewidth}
        \includegraphics[width=\linewidth]{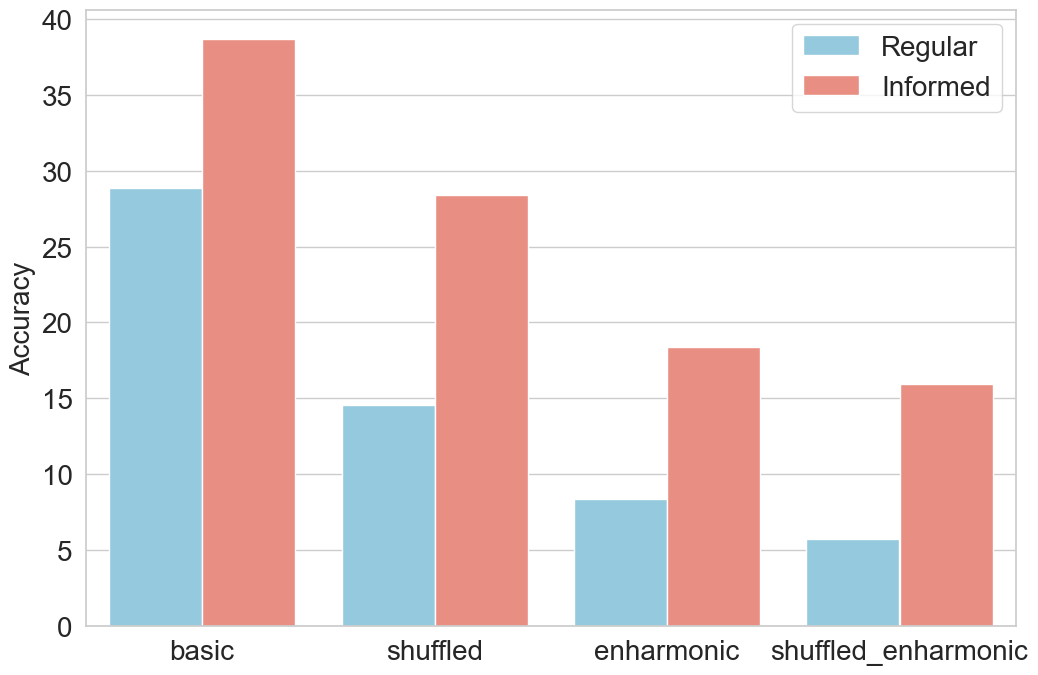}
        \caption{Scales}
        \label{fig:scales}
    \end{subfigure}
    \caption{Results for the task of recognizing chords and scales from the contained notes: First in their basic form only, then with shuffled notes, random enharmonic versions, and both combined. ``Informed'' means that the model was told in advance what chords/scales were possible, and how they had been transformed.}
    \label{fig:res_chords_scales}
\end{figure*}

In the exploration of chords and scales using the LLM, experiments were structured to assess how well the model could recall and apply its knowledge under various configurations. Only GPT-4o was tested in this harder task, where prior knowledge was crucial for success. The results are shown in Figure \ref{fig:res_chords_scales}.

\paragraph{Chords Results}

When the LLM was informed of the chord types, it recognized basic versions fairly accurately. However, without prior information, the model often generated unusual responses. The presence of enharmonic notes significantly reduced accuracy, suggesting that such variations, though known to the model, are not commonly presented in textbooks and hence, are more challenging for the model. Surprisingly, the model performed reasonably well with inversions, although these results were still inferior to the basic chord recognition tasks. In the most complex scenarios, combining multiple variations, the model’s performance dropped dramatically, achieving only about 15\% accuracy without prior information on possible chord types. It was also noted that the model tends to forget which types were previously named, highlighting a possible limitation in its short-term memory.

\paragraph{Scales Results}

Similar trends were observed in the scale identification tasks, with even more pronounced difficulties. The model occasionally attempted to reason through the structures but made errors, such as confusing half and whole steps or misidentifying enharmonic equivalents. Interestingly, despite more information being available for scales, this seemed to confuse the model further rather than aiding it. Even for simpler tasks, such as identifying a scale with seven notes, the model incorrectly named it as a pentatonic scale, indicating a misunderstanding of scale structures.


\section{Conclusion}\label{sec:conclusion}

This study has explored the capabilities of Large Language Models (LLMs) like GPT-3.5 and GPT-4o in understanding and processing musical tasks, particularly focusing on the identification of intervals, chords, and scales. Our results indicate that while LLMs perform adequately in simpler tasks such as identifying note intervals, their performance significantly declines in more complex tasks involving chord and scale recognition.

The experiments revealed that LLMs are heavily influenced by the configurations of the tasks, with changes in interval direction, octave constraints, and accidentals presenting significant challenges. This suggests that LLMs might be relying heavily on patterns observed during training rather than developing a deeper, conceptual understanding of musical theory. The most advanced model, GPT-4o, showed some improvements over GPT-3.5, indicating that model size and training data quality do influence performance. However, the nature of this improvement raises questions about whether the models are truly reasoning through tasks or merely recalling from a broader data set.

Chord and scale tasks demonstrated that without explicit prior information, LLMs often revert to producing unconventional or incorrect answers. Even when models performed better with direct prompting, they frequently failed to apply their knowledge to slightly altered scenarios, such as those involving enharmonic note exchanges or different inversions. These results underscore the limitations of current LLMs in tasks requiring adaptive reasoning and deep understanding. ChatGPT, of course, already contains a module to translate logic problems into code to solve them, which may be helpful here, but does not reflect the reasoning capabilities of the actual model, which are necessary for more sophisticated tasks. 

Our full results, the benchmark data, and its generation code are available under \url{https://github.com/annakaa/harmonic_reasoning}.

\section{Future work}\label{sec:future}

In this work, we focused on reasoning about harmony. Future work could analyze other aspects of music-domain reasoning, e.g. rhythm, or explore non-western music. \cite{yuan2024chatmusician} reported similarly bad results on GPT-4, but showed that In-Context Learning, Chain-of-Thought, and Roleplaying improved results. Furthermore, there are some clues that more sophisticated models demonstrate more emerging reasoning capabilities (see e.g. the differences between GPT-3.5 and GPT-4o).

\bibliography{nlp4MusA}

\newpage
\appendix

\section{Appendix}
\label{sec:appendix}

\begin{table}[ht]
\centering
\caption{Chord and Scale Types}
\begin{tabular}{|l|l|}
\hline
\textbf{Chord Types} & \textbf{Scale Types} \\ \hline
Major & Major \\ 
Minor & Minor \\
Diminished & Dorian \\
Augmented & Phrygian \\
Major Seventh & Lydian \\
Dominant Seventh & Mixolydian \\
Minor Seventh & Locrian \\
Half-Diminished  & Major Pentatonic \\
 Seventh & Minor Pentatonic \\
Fully Diminished & Blues \\
 Seventh & Whole Tone \\
 & Whole-Half Octatonic \\
 & Half-Whole Octatonic \\ \hline
\end{tabular}\label{tab:cs}
\end{table}

\end{document}